\documentclass[10pt,twocolumn,letterpaper]{article}

\usepackage{cvpr}              
\usepackage{epsfig}
\usepackage{amsmath}
\usepackage{amssymb}
\usepackage{graphicx}
\usepackage{tikz}
\usepackage{comment}
\usepackage{color}
\usepackage{pifont}
\usepackage{epsfig}
\usepackage{graphics}
\usepackage{multirow}
\usepackage{algorithm}
\usepackage{algorithmicx}
\usepackage{subcaption}
\usepackage{sidecap}
\usepackage{wrapfig}
\usepackage{enumitem}
\usepackage{colortbl}
\usepackage{arydshln,xcolor,array}
\usepackage{multirow}
\usepackage{esvect}

\usepackage{amsmath}
\usepackage{amssymb}
\usepackage{tikz}
\usepackage{comment}
\usepackage{color}
\usepackage{pifont}
\usepackage{epsfig}
\usepackage{multirow}
\usepackage{algorithm}
\usepackage{algorithmicx}
\usepackage{subcaption}
\usepackage{wrapfig}
\usepackage{enumitem}
\usepackage{colortbl}
\usepackage{arydshln}
\usepackage{xcolor}
\usepackage{array}
\usepackage{esvect}
\usepackage{graphicx}
\usepackage{booktabs}

\newcommand{\model}[0]{GIRAF~}

\newcommand{\mat}[1]{\mathbf{#1}}

\newcommand{\human}{\mat{H}}

\newcommand{\scene}{\mat{S}}

\newcommand{\myparagraph}[1]{\vspace{1pt}\noindent{\textbf{#1}}}

\definecolor{cvprblue}{rgb}{0.21,0.49,0.74}
\usepackage[breaklinks,colorlinks,allcolors=cvprblue]{hyperref}

\def\paperID{3} 
\def\confName{CVPR}
\def\confYear{2026}

\title{GIRAF: Towards Generalizable Human Interactions with Articulated Objects}

\author{
    Xiaohan Zhang$^{1,2,\dagger}$, Sebastian Starke$^{3}$, Alexander Winkler$^{3}$, Federica Bogo$^{3}$, Samir Aroudj$^{3}$,\\ Yuting Ye$^{3}$
}

\makeatletter
\let\@oldmaketitle\@maketitle 
\renewcommand{\@maketitle}{%
    \@oldmaketitle 
    \vspace{-3em} 
    \begin{center}
        \normalsize
        $^1$Tübingen AI Center, University of Tübingen \\
        $^2$Max Planck Institute for Informatics, Saarland Informatics Campus \\
        $^3$Meta Reality Labs Research \\
        \vspace{1em} 
        \includegraphics[width=.82\textwidth]{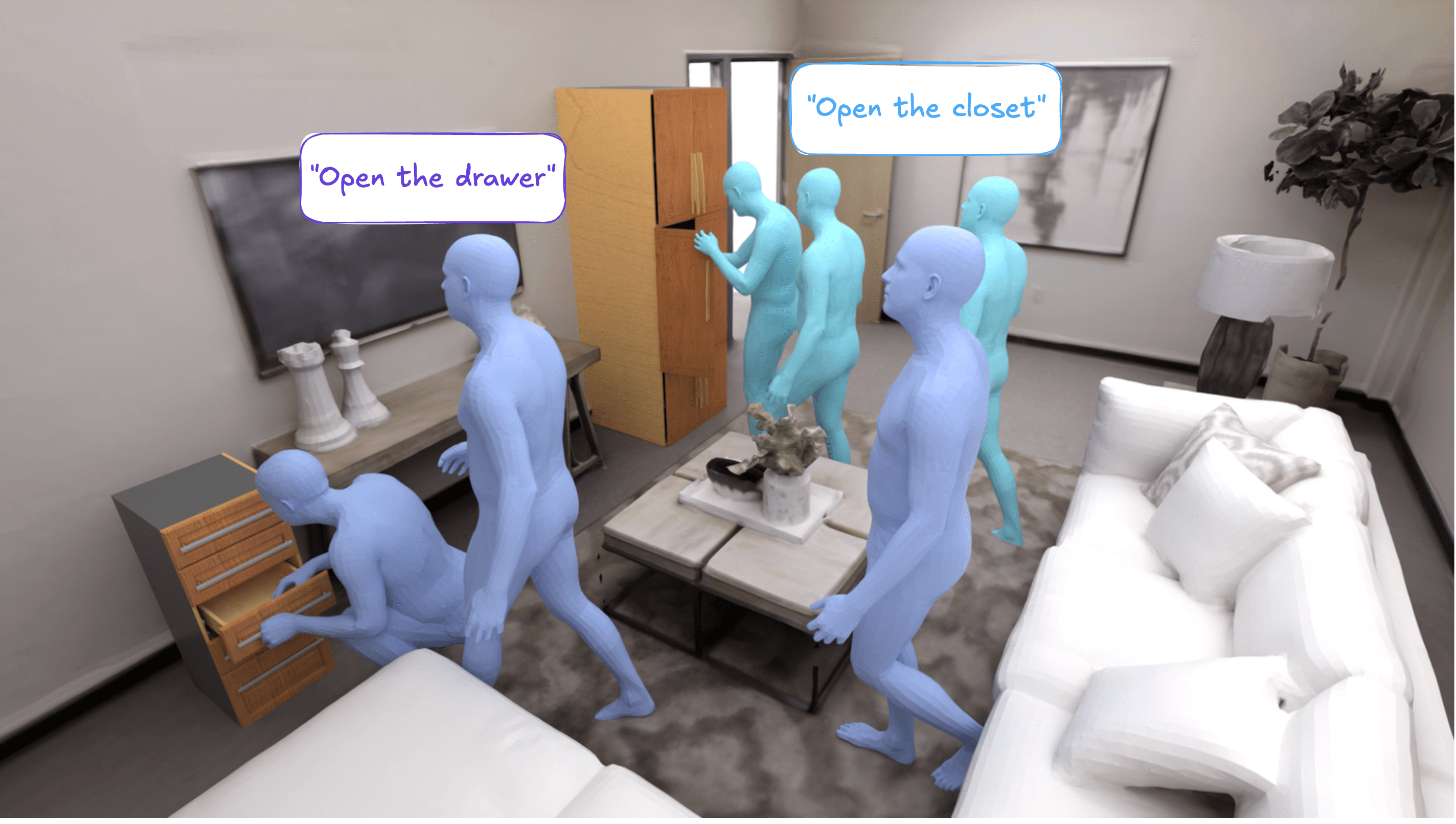}
    \end{center}
    \refstepcounter{figure}
    Figure~\thefigure: Given an initial pose of the human and the object, along with a textual instruction, our goal is to synthesize realistic and physically plausible motion sequences that capture the seamless transition between locomotion and interaction, involving coordinated body, hand, and articulated object motion.
    \label{fig:teaser}
    \bigskip
}
\makeatother

\begin{document}
\maketitle
{\let\thefootnote\relax\footnotetext{$^{\dagger}$Work done during the internship at Meta.}}

\begin{abstract}
Synthesizing realistic full-body human interactions with articulated objects is a fundamental challenge for embodied AI and graphics, with applications in robotics training and virtual agents. Existing models remain limited: some focus on simple activities with static objects, while others restrict attention to hand-only manipulation. This leaves open the problem of generating coordinated full-body motion that approaches, manipulates, and moves articulated objects in a realistic and generalizable way. The key difficulty lies in reasoning jointly about locomotion, fine-grained contact, and object articulation. Models must capture subtle hand--object correspondences that transfer across object geometries, while also producing seamless transitions from navigation to manipulation. At the same time, the scarcity of large-scale paired motion--scene data makes it difficult to generalize across diverse object positions and shapes. We introduce a text-conditioned diffusion model that addresses these challenges through three core ideas: an object-centric representation that unifies hand--object contact with object surfaces, a mixed-domain training strategy that balances locomotion and interaction, and a contact-based augmentation scheme that expands training diversity. Through experiments, our method demonstrated strong generalization to unseen object configurations, surpassing current state-of-the-art methods.
\end{abstract}

\section{Introduction}

Generating realistic human motion while interacting with objects has broad applications in virtual reality and in embodied AI and robotics, where it enhances immersion and provides a scalable source of synthetic data to train agents that must interact with the physical world. While significant progress has been made, existing methods focus on narrow settings: some address simple full-body activities such as sitting on a chair, while others emphasize dexterous hand-only interactions such as grasping a cup. In this work, we present a unified approach that enables full-body interactions with articulated objects, for example ``opening a refrigerator and take something out'' or ``pulling out a drawer''. Given an initial pose of the human and the object, along with a textual instruction, our goal is to synthesize realistic and physically plausible sequences that capture the seamless transition between locomotion and interaction, involving coordinated body, hand, and articulated object motion.

Achieving this goal is challenging for several reasons. First, modeling fine-grained hand--object contact is difficult. Previous methods either define contact on object surfaces, which fails to generalize to new shapes, or predict contact on the hand, which does not ensure consistent correspondence. Second, a model must handle the transition from locomotion, where the human approaches the object, to interaction, where manipulation takes place. This requires reasoning across different motion regimes, yet prior works typically assume the human is already in contact or within close vicinity, leaving locomotion unaddressed. Third, generalization to new object configurations is critical. In real-world settings, articulated objects vary in shape, scale, and position, but many existing methods only evaluate on placements seen during training, limiting robustness. Finally, acquiring large-scale paired motion--scene datasets is costly and impractical due to the difficulty of capturing motion in cluttered environments, which creates a data scarcity bottleneck.

To address these challenges, we propose a text-conditioned diffusion model for synthesizing long-horizon full-body interactions with articulated objects. First, to handle fine-grained interaction, we unify hand--object contact, hand end-effectors, and object surfaces into a shared object-centric representation, which avoids the pitfalls of hand- or object-only encodings. Second, to enable seamless transitions between locomotion and interaction, we design a mixed-domain training strategy with homogeneous batches and an annealing schedule, which ensures balanced learning across different motion types. Third, we use FiLM-based conditioning to preserve interaction-specific cues while still allowing the model to represent locomotion naturally. Finally, to improve generalization under limited data, we propose a contact-based data augmentation strategy that relocates interactions while maintaining contextual plausibility.

Together, these advances yield a unified text-conditioned diffusion framework for generating realistic full-body locomotion and articulated-object interaction sequences. Our experiments show that the model produces physically plausible, long-term motion across diverse environments and generalizes effectively to unseen object placements. Through experiments,~\model surpasses current baselines quantitatively and qualitatively.

The contributions of this work are threefold:
\begin{enumerate}
\item We introduce a text-conditioned diffusion model for unified full-body locomotion and interaction with articulated objects.
\item We propose an object-centric contact representation that enables fine-grained and generalizable hand--object interactions.
\item We develop a robust training strategy that combines mixed-domain learning, FiLM-based conditioning, and contact-driven data augmentation to support long-horizon motion synthesis in complex 3D environments.
\end{enumerate}

\section{Related Work}
\paragraph{Environment-Aware Human Motion Synthesis.}
Recent years have seen growing interest in synthesizing human motion conditioned on 3D environments. Early work focused on interactions with static objects such as chairs and beds~\cite{zhang2022couch,samp,zhang2023roam,jiang2023chairs,nsm,Pi_2023_ICCV,pan2023synthesizing,yi2024tesmo,kulkarni2023nifty}, while later approaches extended to dynamic objects~\cite{kim2024parahome,zhang2024hoi,bhatnagar22behave,zhang2024core4d}. These models often assume continuous contact and thus overlook the transition from locomotion to manipulation~\cite{THOR,xu2023interdiff,diller2023cghoi,xu2024interdreamer,peng2023hoi,yang2024fhoi,xu2025interact,xu2024interdreamer}. Other works integrate navigation and locomotion~\cite{zhang2024force,li2023controllable,li2023OMOMO,yi2024tesmo,li2024chois,trumans}, or focus on terrain traversal and scene navigation~\cite{pfnn,cong2024laserhuman,zhang2024scenic}, but generally neglect hand motion.

Following the recent emergence of powerful diffusion models for motion~\cite{tevet2023human,shafir2023human,zhang2022motiondiffuse,zhang2023remodiffuse,chen2023executing,dabral2022mofusion,HoangGGM24,humantomato,ma2024richcat,zhang2024tedi,zhou2023emdm,KongGLMW23}, more recent work incorporates text control into human-scene interaction. TRUMANS~\cite{trumans} unified static and dynamic object interactions. However, these models still assume flat terrains or floors. While some concurrent works have demonstrated human motion on stairs~\cite{Zhao:DART:2024,cong2024laserhuman}, they have the limitation of not training on paired motion-scene data. This lack of scene awareness restricts the model's ability to generalize to complex terrain surfaces. Moreover, their approach requires the future 3D root position, which is not always available.

In parallel, research on dexterous manipulation studies fine-grained hand interactions with small objects using datasets such as GRAB~\cite{GRAB:2020} and ARCTIC~\cite{fan2023arctic}. These methods~\cite{liu2024geneoh,christen2022dgrasp,taheri2023grip,manipnet,braun2023physically,dfbgrasp2024braun,Zheng_2023_CVPR,christen2024diffh2o,cha2024text2hoi,huang_etal_cvpr25,taheri2024grip} achieve impressive fidelity but are restricted to local hand-centric motion without considering whole-body dynamics. In broader scene interaction, most approaches remain short-term~\cite{wang2022humanise,cen2024text_scene_motion,wang2024move} or depend on strong control signals like target poses~\cite{Zhao:ICCV:2023,liu2023mob} or simplified geometry assumptions~\cite{trumans,mir23origin,lee2023lama}.
A separate line of work leverages reinforcement learning, either distilling pretrained priors for task-specific skills~\cite{Luo2023PerpetualHC,tevet2025closd,uniphys,luo2024universal} or training agents for interaction~\cite{hassan2023,merel2020catchcarry,cui2024anyskill,tessler2024maskedmimic,xu2025intermimic} and terrain traversal~\cite{luo2024universal,rempeluo2023tracepace}. While physically grounded, these methods require extensive task-specific design and often lack human-like realism.

In contrast, our work unifies locomotion, hand--object contact, and object articulation in a text-driven diffusion framework, enabling the generation of long-horizon, realistic full-body interactions with articulated objects in complex environments.

\subsection{Full-Body Human--Object Interaction}

Synthesizing full-body human--object interactions is inherently challenging as it requires coordinated human motion and physically plausible object dynamics. Early work explored static full-body grasps~\cite{tendulkar2022flex}, or generated grasping motions while neglecting object movement~\cite{taheri2021goal}. IMoS~\cite{ghosh2022imos} generates body motion auto-regressively while optimizing object trajectories under the assumption of a fixed hand--object contact frame. TOHO~\cite{toho} similarly synthesizes full-body motion via implicit representations, but recovers object dynamics by relying on the same static-contact assumption. DiffGrasp~\cite{diffgrasp} introduced diffusion models for whole-body interaction, conditioned on pre-specified object trajectories with hand--object guidance to refine contact quality. Other physics-based approaches~\cite{wang2023physhoi,wang2024skillmimic,xiao2025motionstreamer} track manipulation behaviors but produce short and often unnatural motions. More recent extensions tackle humanoid grasping~\cite{omnigrasp,li2025walkingref}, yet remain limited to rigid objects and simple contacts. Crucially, articulated object manipulation is far more complex: it requires reasoning about part-specific contact regions and generating motion that actuates the articulation itself.

Recent methods begin to move towards this direction. HOIDiNi~\cite{ron2025hoidini} adapts a pre-trained diffusion prior through noise optimization to produce full-body interactions with small GRAB~\cite{GRAB:2020} objects, but its reliance on directly predicted contact points on the object restricts generalization to unseen categories, and it does not address articulated objects. CoDA~\cite{pi2025coda} extends diffusion models to articulated settings using three separate networks for body, hand, and object motion, but depends on ambiguous end-effector positions from a predicted distance field and requires expensive optimization. Later work~\cite{lingo} replaces action labels with free-form text prompts, yet contact modeling is limited to a few fingertip keypoints, and object articulation remains unsupported. HOIFHLI~\cite{wu2024human} leverages LLM-based scene analysis and nearest-neighbor hand contact projection, but is restricted to rigid objects.

In contrast, our approach tackles the problem of full-body articulated object interaction directly. We introduce a unified diffusion framework that synthesizes long-horizon motion where the human approaches, manipulates, and actuates articulated objects, while reasoning jointly about fine-grained contact, object geometry, and locomotion. Unlike prior work, our method generalizes across object configurations and supports seamless transitions between navigation and manipulation within a single model.

\section{Data Representation}
In this section, we define the inputs and outputs considered in this work. Given an initial pose of the human and the object, along with a textual instruction, our goal is to synthesize realistic and physically plausible motion sequences that capture the seamless transition between locomotion and interaction, involving coordinated body, hand, and articulated object motion.

\myparagraph{Object representation.}
We consider objects from ParaHome~\cite{kim2024parahome}, which include two-part articulated objects with two degrees of freedom. Each object state is represented as $\scene_{o} = \{R_{o}, a_{o}\}$, where $R_{o}$ denotes the hinged rotation angle and $a_{o}$ denotes the translation of the movable part of the object.

\myparagraph{Human representation.}
We represent the human body using SMPL-X~\cite{SMPL-X:2019}, a parametric 3D body model that includes the face, hands, and fingers. SMPL-X is a differentiable function that takes shape, pose, and expression parameters as input and outputs a mesh with 10,475 vertices and 20,908 faces, posed using linear blend skinning with a rigged skeleton. Since our focus is on body and hand motion, we remove the face-related parameters. Formally, the human representation $\human = \{H\}_{t=1}^{L}$ has the following components $\{\Theta, \gamma, \tilde{J}_{\text{hands}}\}$. Let $\Theta = \{\theta, t\}$ denote the SMPL-X pose parameters, where $\theta \in \mathbb{R}^{52 \times 6}$ are the joint rotations in 6D rotation representations~\cite{zhou_6d} and $\gamma\in \mathbb{R}^{3}$ is the global root translation.

\myparagraph{Text Embedding.}
Each motion sequence in ParaHome~\cite{kim2024parahome} is segmented and annotated with a textual description of the object being manipulated and the corresponding human action, such as \emph{``a person opening the microwave''} or \emph{``a person taking something out of the drawer''}. We encode the textual descriptions using a pre-trained DistilBERT model~\cite{sanh2019distilbert}, which provides a compact semantic embedding as input to our framework.

\section{Method}

\begin{figure*}[ht!]
    \centering
    \includegraphics[width=0.90\textwidth]{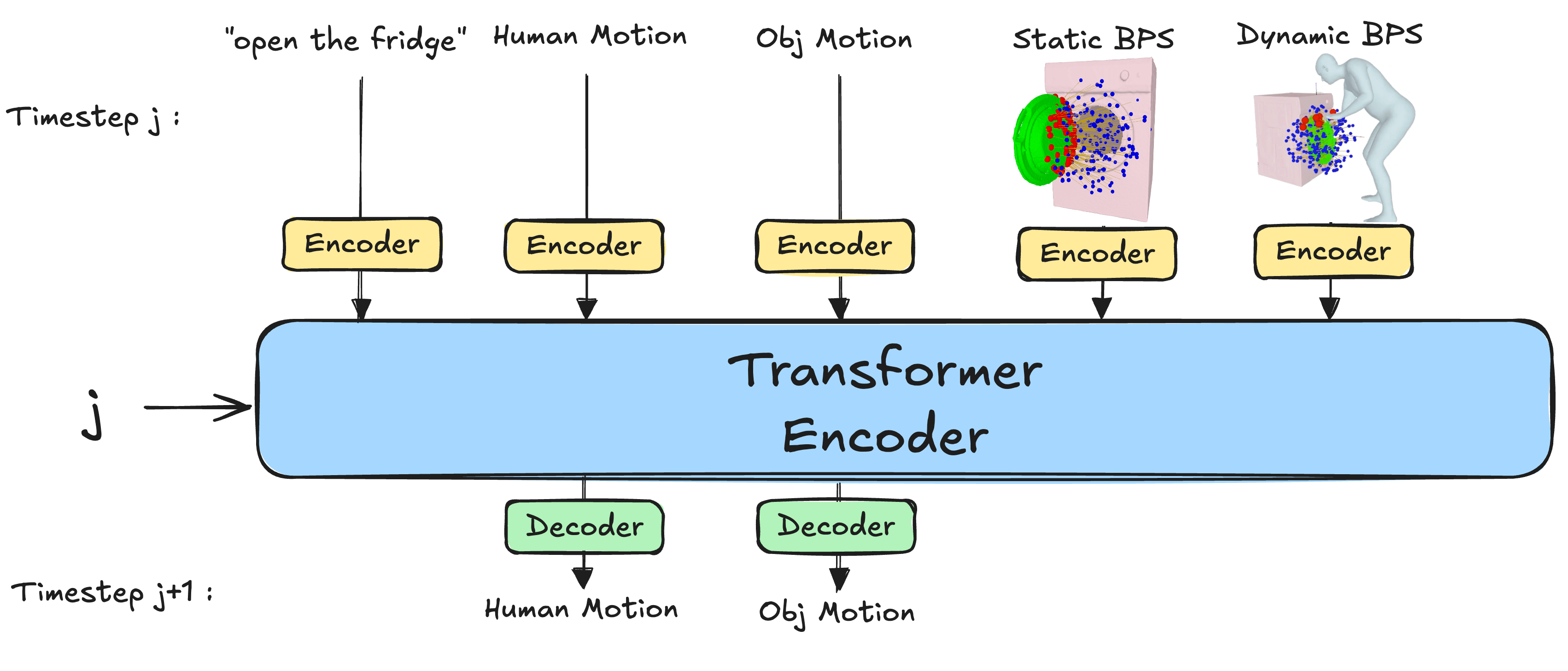}
    \caption{Architecture overview.~\model leverages a transformer-based diffusion model. Given a current pose of the human and the object, a textual instruction and the human-object interaction representations, our method synthesizes realistic and physically plausible future human and object motion.}
    \label{fig:arch}
\end{figure*}

Our goal is to synthesize realistic and physically plausible sequences of human motion interacting with articulated objects, conditioned on text instructions and initial object states. Figure~\ref{fig:arch} provides an overview of our framework. At the core is a text-conditioned diffusion model, which must simultaneously reason about fine-grained hand-object contact, seamless transitions between locomotion and interaction, and generalization to new object placements. To address these challenges, we introduce four complementary components: a novel object-centric representation of hand-object interaction, a mixed-domain training strategy, a contact-based augmentation scheme, and scene-aware diffusion noise optimization. Together, these components form a coherent pipeline that enables long-horizon, generalizable human-object interaction synthesis.

\subsection{Dynamic BPS: Unifying Contact, Hand and Object}
A central challenge is representing fine-grained contact in a way that generalizes across diverse object geometries. Prior approaches either encode contact on object surfaces, which fails to transfer to new shapes, or predict contact on hand keypoints, which does not guarantee consistent correspondence on the object surface.

To overcome this, we propose \textit{dynamic basis point sets (BPS)} canonicalized to the articulated part of the object (e.g., the door of a microwave or the moving part of a drawer). A basis point set is defined by a fixed set of points $P \in \mathbb{R}^{K \times 3}$ sampled within the unit sphere centered at the object origin. The object mesh is normalized to unit scale, and distances from $P$ to the object surface are computed as features.

Dynamic BPS integrates three features in the same object-centric space: (1) object distances $o_{d} \in \mathbb{R}^K$, (2) distances to hand end-effectors $\tilde{J}_{\text{hand}}^t \in \mathbb{R}^{12 \times K}$, and (3) binary contact labels $c^t \in \{0,1\}^K$. A basis point is considered in contact if its corresponding surface point lies in contact with a hand joint. This design provides two key advantages. First, it is shape-agnostic: contacts are represented through a voting scheme over shared basis points, which encourages generalization. Second, it can be combined with hand-centric features for improved accuracy, enabling more precise correspondence between hand and object surfaces.

\subsection{Mixed-Domain Training for Locomotion and Interaction}
While the object-centric representation enables reasoning about contact, the model must also handle the transition between locomotion and interaction. Humans first approach an object and then manipulate it, requiring the model to reason across distinct motion regimes. Prior work often bypasses this complexity by assuming the human is already near the object.

We introduce a mixed-domain training strategy that balances locomotion and interaction sequences. Directly zeroing features for locomotion introduces ambiguity, especially for distance features in object BPS space. Instead, we integrate FiLM layers~\cite{perez2018film} with trainable embeddings into our transformer blocks. These embeddings preserve interaction-specific features while allowing the model to represent locomotion naturally.

During training, we apply a locomotion mask $M_{\text{loco}} \in [0,1]^L$ over a sequence of length $L$, marking frames where the human is more than 0.5 meters from the object. This supervision allows the model to learn smooth transitions between approaching and interacting behaviors. Furthermore, we adopt a batch mixing strategy: early training uses balanced batches of locomotion and interaction, while later training anneals toward interaction-heavy batches. This annealing schedule improves the synthesis of long-horizon behaviors without sacrificing locomotion fidelity. It is also found that training with homogeneous batches is imperative for model performance.

\begin{table*}[ht!]
    \small
    \centering
    \caption{Contact-based and penetration evaluation.}
    \resizebox{\textwidth}{!}{
        \begin{tabular}{lccccccc}
            \hline
            Method & Contact Dist (cm)$^{\downarrow}$ & Contact Precision$^{\uparrow}$ & Contact Recall$^{\uparrow}$ & Contact Accuracy$^{\uparrow}$ & Contact F1$^{\uparrow}$ & Penetr. Dist (cm)$^{\downarrow}$ & Penetration $^{\downarrow}$ \\
            \hline
            LINGO~\cite{lingo}  & 2.502  & 0.7673  & 0.7769  & 0.7193  & 0.7416  & 1.094  & 0.4736  \\
            CHOIS~\cite{li2024chois}   & 2.288  & 0.7848  & 0.7874  & 0.7504  & 0.7577  & 1.071  & 0.4781  \\
            \hline
            Ours   & \textbf{1.869}  & \textbf{0.7824}  & \textbf{0.8420}  & \textbf{0.7800}  & \textbf{0.7898}  & \textbf{1.044}  & \textbf{0.4572}  \\
            \hline
        \end{tabular}
    }
    \label{table:quant_contact}
\end{table*}

\begin{figure*}[ht]
    \centering
    \includegraphics[width=\textwidth]{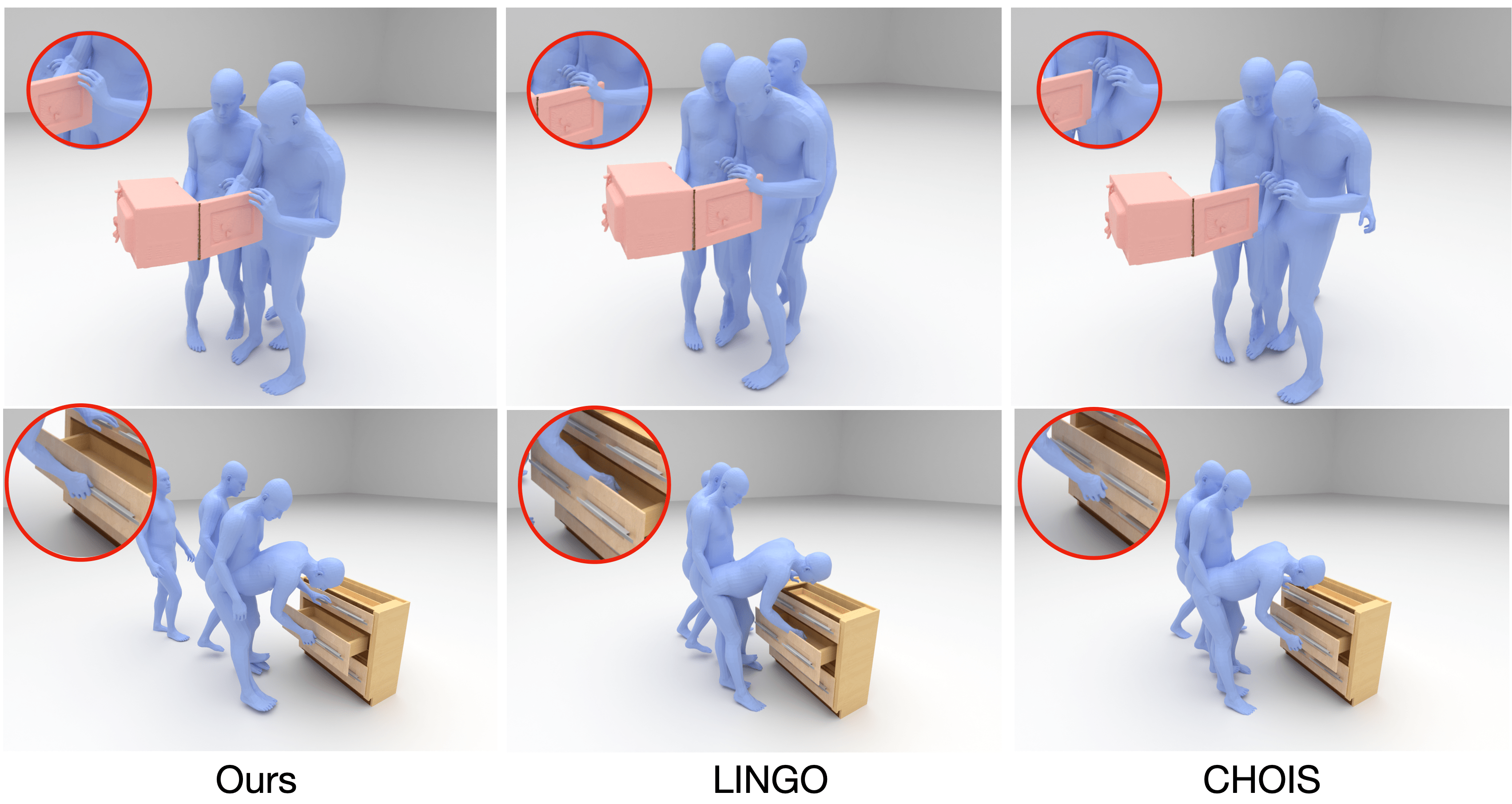}
    \caption{Qualitative comparison with baselines. By comparing with the baselines, it can be shown that with our novel human-object interaction representation, our model is able to synthesize more realistic and physically plausible human-object interactions.}
    \label{fig:qualitative}
\end{figure*}

\subsection{Contact-Based Augmentation for Generalization}
Even with a unified representation and mixed-domain training, data scarcity remains a major obstacle. Collecting paired human-object motion data across diverse layouts is costly and does not scale. To improve robustness, we introduce a contact-based augmentation strategy that relocates objects while preserving the plausibility of interactions.

We first identify hand-object contact points in the original sequence, then map them to corresponding points on the transformed object. This ensures that the contact relationship is preserved even when the object is resized or relocated. To achieve this, we place objects on a discretized 3D grid spanning a $0.1 \times 0.2 \times 0.7$ meter space with resolution $0.1$m. Additionally, we randomly rescale the object to generalize to different shapes. Human motion is recomputed using a CCD-based inverse kinematics solver, with rotational limits applied to maintain natural articulation. This augmentation allows the model to experience diverse object placements during training, significantly improving generalization at test time.

\subsection{Scene-Aware Diffusion Noise Optimization}
After training, the diffusion model is capable of generating coordinated locomotion and interaction. However, naive sampling may still produce minor artifacts such as hand-object penetration or jittery motions. To refine generation quality, we adopt scene-aware diffusion noise optimization~\cite{karunratanakul2023dno}. We adopt DDIM sampling~\cite{song2020ddim} to efficiently generate motion sequences during optimization. The objectives are computed on the final output, and gradients are propagated back through the DDIM solver to update the noise. After optimization, we pass the optimized noise into the diffusion models to generate the final whole-body motion.

In the first stage, we align object motion with predicted hand contacts by optimizing diffusion noise against a contact loss
\[
\mathcal{L}_{\text{contact}} = \frac{1}{|v_o|}\sum_{v \in v_o} \|v - J_{\text{hands}}\|^2,
\]
where $v_o$ are the object vertices and $J_{\text{hands}}$ are the hand joint positions. In the second stage, we reduce hand-object penetration with a collision loss
\[
\mathcal{L}_{\text{collision}} = \text{SDF}(\mathbf{v}),
\]
where $\mathbf{v}$ are body mesh vertices and SDF is the signed distance field of the object mesh. Finally, to suppress jitter, we apply a smoothness loss
\[
\mathcal{L}_{\text{smooth}} = \left\|\mathbf{J}_p^{1:N} - \mathbf{J}_p^{0:N-1}\right\|_2,
\]
where $\mathbf{J}_p$ are global joint positions over time. This two-stage optimization significantly improves realism, yielding fluid and physically consistent motion.

\section{Experiments}

In this section, we present a comprehensive evaluation of our approach. We begin by introducing the datasets and implementation details. We then compare our method against strong baselines on multiple quantitative metrics covering contact, penetration, pose accuracy, and text-to-motion alignment. We also provide qualitative comparisons and additional generalization results in later subsections.

\begin{figure*}[ht]
    \centering
    \includegraphics[width=\textwidth]{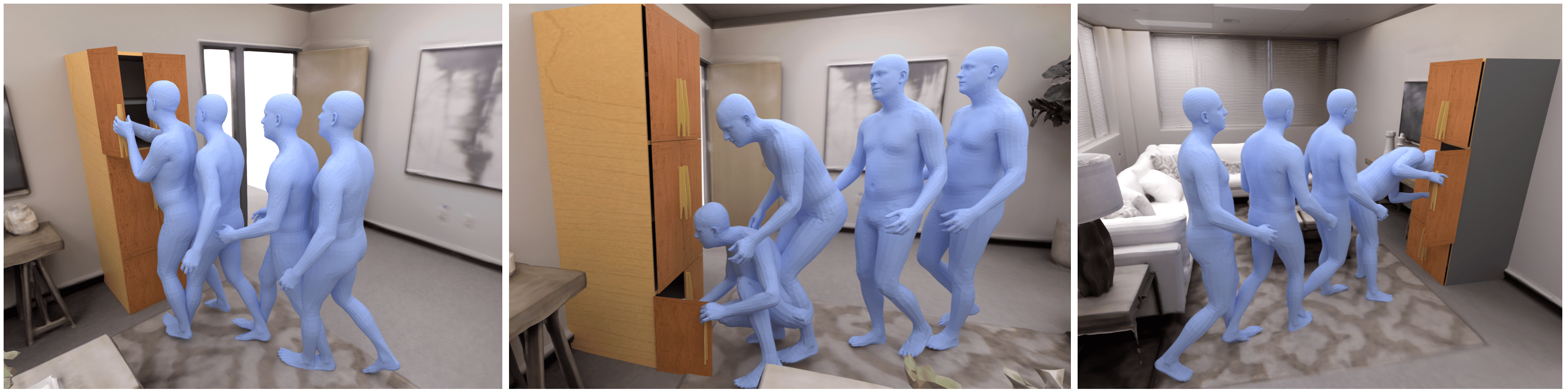}
    \caption{Our model synthesizes realistic sequences of opening different doors of a closet, despite these configurations not being seen during training. The human adapts body motion, reach, and hand placement to each door's geometry, demonstrating that our unified object-centric representation enables generalization beyond fixed placements.}
    \label{fig:closet}
\end{figure*}

\begin{figure*}[ht!]
    \centering
    \includegraphics[width=\textwidth]{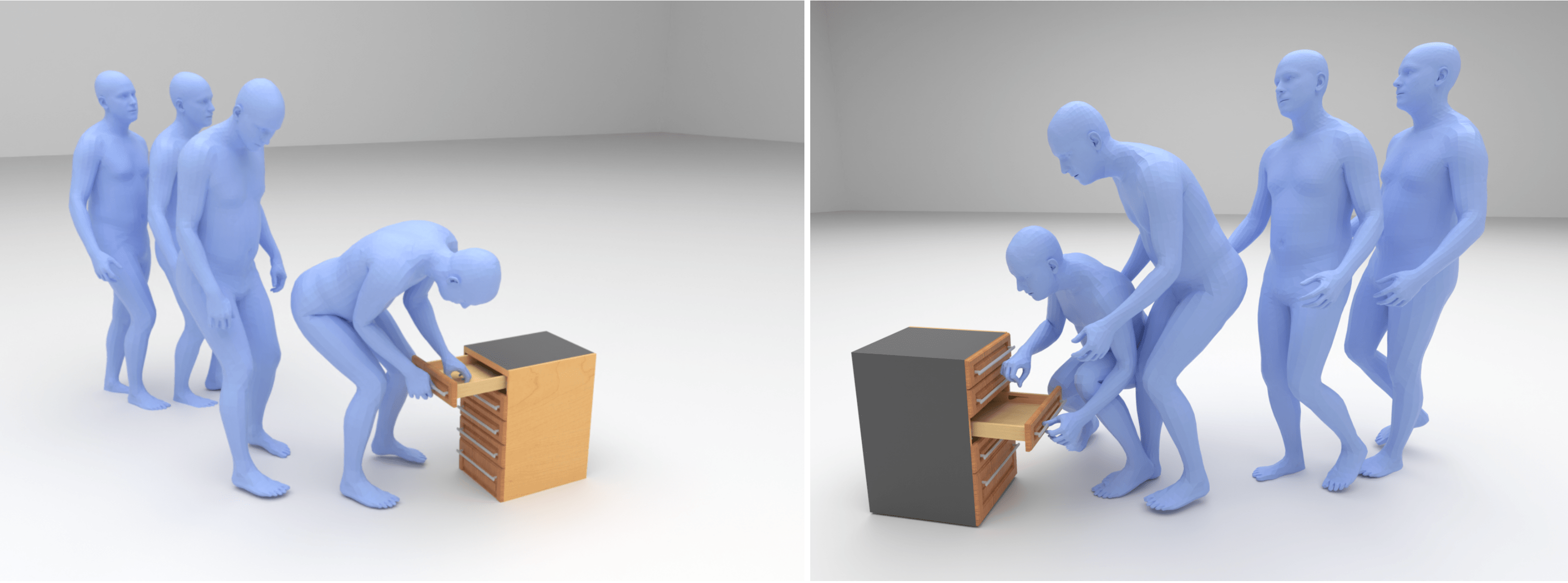}
    \caption{We show synthesized sequences of ``open the drawer'' at two different drawer heights. The model produces consistent reaching and pulling actions adapted to the drawer position, preserving both contact quality and whole-body coordination. This highlights the robustness of our approach to spatial variation within the same object category.}
    \label{fig:drawer}
\end{figure*}

\subsection{Dataset}
For training, we leverage the ParaHome~\cite{kim2024parahome} dataset, which contains articulated household objects paired with annotated interaction descriptions. Each motion sequence is segmented and labeled with natural language phrases describing the human action and the object, e.g., \emph{``a person opening the microwave''} or \emph{``a person taking something out of the drawer''}. From this dataset, we extract four interaction categories: \emph{drawer}, \emph{microwave}, \emph{refrigerator}, and \emph{washing machine}, yielding a total of 1.5 hours of interaction data.

For locomotion, we augment the dataset with segments from Babel~\cite{BABEL:CVPR:2021}, retrieving clips annotated with locomotion categories such as \emph{walk}, \emph{run}, \emph{jog}, \emph{step}, and \emph{crouch}. This combination allows us to train a model capable of handling both free-form locomotion and fine-grained object interaction. After applying our contact-based augmentation strategy, the final dataset contains 2100 sequences spanning 3 hours of motion.

\begin{table}[ht!]
    \small
    \caption{Quantitative comparison on pose and object reconstruction errors.}
    \centering
    \resizebox{\columnwidth}{!}{
        \begin{tabular}{lcccc}
            \hline
            Method & MPJPE$^{\downarrow}$ & MPVPE$^{\downarrow}$ & Hand Error$^{\downarrow}$ & Obj. Error$^{\downarrow}$ \\
            \hline
            LINGO~\cite{lingo}  & 0.1376  & 0.1836  & 0.1587  & 0.0578  \\
            CHOIS~\cite{li2024chois}  & 0.1297  & 0.1757  & 0.1537  & 0.0556  \\
            \hline
            Ours   & \textbf{0.1143}  & \textbf{0.1704}  & \textbf{0.1328}  & \textbf{0.0513}  \\
            \hline
        \end{tabular}
    }
    \label{table:quant_pose}
\end{table}

\begin{figure*}[ht!]
    \centering
    \includegraphics[width=\textwidth]{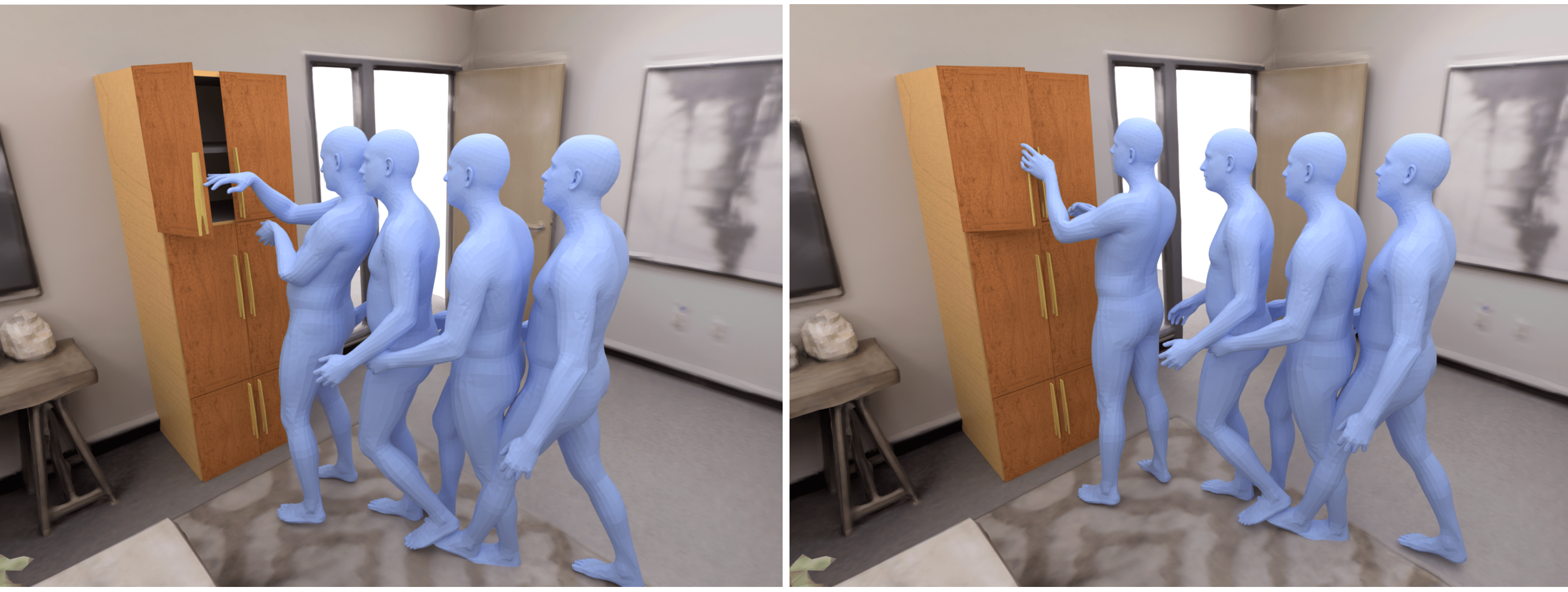}
    \caption{For the same object and instruction, our model can synthesize interactions with different hand contact strategies, such as using the left hand, right hand, or both hands. This controllability stems from our text-conditioned design, which allows semantic variation in contact strategies while maintaining physically plausible interaction and consistent locomotion.}
    \label{fig:contact}
\end{figure*}

\subsection{Implementation Details}
Our diffusion backbone is based on a transformer encoder with 8 layers and a latent dimension of 256. The trainable embeddings used in FiLM layers for locomotion are initialized from a Gaussian distribution $\mathcal{N}(0, 0.02)$. We train the model with Adam optimizer using a cosine learning rate schedule, decaying from $1\mathrm{e}{-4}$ to $1\mathrm{e}{-5}$, over 30,000 epochs with a batch size of 64, taking up to 1 day on a single A100 GPU. During training, we apply our mixed-domain strategy with linear annealing between locomotion- and interaction-focused batches. At inference time, we run 100 diffusion steps for all models, ensuring consistent comparison against baselines. For evaluation, each method is conditioned on the exact initial human pose, object state, and text description from the test sequence, and reconstruction-style metrics are computed frame-wise against the paired ground-truth continuation.

\subsection{Baselines}
We train all baselines on the same \model dataset. LINGO~\cite{lingo} and CHOIS~\cite{li2024chois} have achieved impressive results on human-object interaction given textual instructions. Since neither of the methods considers hand motion, we incorporate hand motion into the baseline model training.

\subsection{Quantitative Evaluation}
We evaluate our method against two baselines, LINGO~\cite{lingo} and CHOIS~\cite{li2024chois}, across three groups of metrics: (1) contact and penetration quality, (2) text-to-motion alignment, and (3) pose and object reconstruction accuracy. We utilize well-established quantitative metrics following human-object interaction~\cite{Zhao:DartControl:2025,li2024chois,li2023OMOMO,wu2024human} and text-to-motion literature~\cite{tevet2023human,shafir2023human,zhang2022motiondiffuse,zhang2023remodiffuse,chen2023executing,dabral2022mofusion,HoangGGM24,humantomato,ma2024richcat,zhang2024tedi,zhou2023emdm,KongGLMW23}.

\myparagraph{Contact and penetration.}
Table~\ref{table:quant_contact} reports quantitative measures of physical plausibility. We use \textbf{contact distance} (cm) to assess the average distance between predicted hand joints and object surfaces, along with \textbf{precision}, \textbf{recall}, \textbf{accuracy}, and \textbf{F1 score} for contact prediction. Penetration is evaluated both by mean distance (cm) and frequency. Our method achieves the lowest contact distance (1.869 cm) and penetration error (1.044 cm), while improving both precision and recall, highlighting its ability to produce realistic, physically consistent hand-object interactions.

\myparagraph{Text-to-motion quality.}
To assess semantic alignment, we report results in Table~\ref{table:quant_quality} using four metrics: \textbf{T2M retrieval} (lower is better), \textbf{R-precision}, \textbf{Frechet Inception Distance} (FID), and \textbf{diversity}. Our method consistently outperforms baselines, achieving the best R-precision (0.3812) and the lowest FID (0.0634), while maintaining high diversity (3.082). These results demonstrate that our diffusion model not only respects the text descriptions but also produces motions that remain varied and natural.

\myparagraph{Pose and object accuracy.}
Table~\ref{table:quant_pose} evaluates reconstruction accuracy with mean per-joint position error (MPJPE), mean per-vertex position error (MPVPE), mean hand error, and object error. Our method again achieves the lowest error across all categories, reducing MPJPE to 0.1143 and hand error to 0.1328. This indicates more precise articulation of both body and hands, as well as better synchronization with the manipulated objects.
These reductions mean the predicted hand trajectories stay closer
to the object surfaces while avoiding implausible interpenetration.

\begin{table}[ht!]
    \small
    \centering
    \caption{Quantitative comparison on text-to-motion quality metrics.}
    \resizebox{\columnwidth}{!}{
        \begin{tabular}{lcccc}
            \hline
            Method & T2M$^{\rightarrow}$ & R-precision$^{\uparrow}$ & FID$^{\downarrow}$ & Diversity$^{\uparrow}$ \\
            \hline
            GT     & 0.4514 & 0.3691 & 0      & 2.968 \\
            LINGO~\cite{lingo}  & 0.6076 & 0.3563 & 0.0697 & 2.723 \\
            CHOIS~\cite{li2024chois}  & 0.5830 & 0.3750 & 0.0791 & 2.892 \\
            \hline
            Ours   & \textbf{0.5115} & \textbf{0.3812} & \textbf{0.0634} & \textbf{3.082} \\
            \hline
        \end{tabular}
    }
    \label{table:quant_quality}
\end{table}

\subsection{Qualitative Comparison}
As shown in Figure~\ref{fig:qualitative}, we present side-by-side qualitative comparisons between our method and two strong baselines, LINGO~\cite{lingo} and CHOIS~\cite{li2024chois}. Each set depicts three results generated from the same textual instruction. While baselines can produce plausible global body motion, they often fail to resolve fine-grained hand--object interactions: hands either float near the surface without contact or penetrate deeply into the object. This is largely due to their lack of explicit reasoning about hand, object, and contact geometry in a unified space.

By contrast, our approach introduces a novel dynamic object-centric BPS representation that encodes body, hands, and articulated objects in the same space. This unified encoding allows our model to generate sequences that exhibit coherent hand placements, physically consistent contact, and minimal penetration. Notably, our results maintain contact across the interaction horizon (e.g., pulling a drawer or opening a microwave) while adapting the body pose naturally, which cannot be achieved by methods that treat hands and objects separately. These qualitative comparisons validate the quantitative improvements reported in Tables~\ref{table:quant_contact}--\ref{table:quant_pose}, highlighting the importance of modeling interaction in a shared dynamic representation.

\subsection{Generalization and Controllability}
Beyond producing realistic motion for the training categories, our model generalizes robustly to new object configurations and supports controllable generation, as illustrated in Figures~\ref{fig:closet}--\ref{fig:contact}. In Figure~\ref{fig:closet}, the human is instructed to open different doors of a closet. By conditioning on the articulated part geometries, our model adapts seamlessly to novel configurations without retraining, generating plausible reach, grasp, and actuation sequences. Similarly, in Figure~\ref{fig:drawer}, the same action \emph{``open the drawer''} is synthesized across different drawer heights, showing that the model generalizes to varying placements of the same object type.

Finally, Figure~\ref{fig:contact} demonstrates controllability at the level of hand contact. Given the same object, our model can be directed via text to use one hand, both hands, or alternative contact strategies. This flexibility arises from our text-conditioning design, which allows interaction semantics to be specified without sacrificing physical plausibility. Together, these results show that our method is not only accurate on standard benchmarks but also versatile in real-world scenarios where objects appear in diverse configurations and humans adopt different strategies to interact with them.

\section{Conclusion}
We presented a diffusion-based framework for generating realistic full-body human interactions with articulated objects, conditioned on text instructions. Our method tackles three fundamental challenges: reasoning about fine-grained hand--object contact through an object-centric representation, enabling seamless transitions between locomotion and interaction via a mixed-domain training strategy, and improving generalization through contact-based augmentation. Together, these components allow our model to synthesize long, physically plausible motions that respect object articulation and adapt to diverse environments and textual descriptions. Experiments on ParaHome show that our approach improves over current baselines for long-horizon locomotion and articulated-object interaction.

While our model shows strong generalization within the ParaHome domain, it remains limited in handling novel articulated mechanisms not observed during training, such as doors with rotational joints. We also observe occasional motion artifacts, including brief foot sliding or local joint jitter near locomotion-to-interaction transitions, which likely stem from the difficulty of jointly modeling mixed locomotion and manipulation regimes under limited paired training data. Moreover, the current framework focuses on locomotion and short-range manipulation in simple terrains, and does not yet support more complex navigation scenarios like stair traversal with interaction. Future work will explore extending our representation to cover a broader set of articulated object categories, as well as integrating more diverse locomotion behaviors into a single compact model. We believe these directions will broaden applicability in robotics and virtual agents.

{
    \small
    \bibliographystyle{ieeenat_fullname}
    \bibliography{main}
}


\end{document}